\pdfoutput=1

\documentclass[11pt]{article}

\usepackage[]{acl}

\usepackage{times}
\usepackage{latexsym}

\usepackage[T1]{fontenc}

\usepackage[utf8]{inputenc}

\usepackage{microtype}

\usepackage{smile}
\usepackage{hyperref}
\usepackage{url}
\usepackage{booktabs}
\usepackage{makecell}

\newcommand{\ours}{{CAMERO}}

\title{CAMERO: Consistency Regularized Ensemble of Perturbed Language Models with Weight Sharing}

\author{Chen Liang \thanks{~~Work was done during an internship at Microsoft Azure AI.} \\
  Georgia Institute of Technology \\
  \texttt{cliang73@gatech.edu} \And
  Pengcheng He, Yelong Shen \\
  Microsoft Azure AI \\ 
  \texttt{\{penhe,yelong.shen\}@microsoft.com} \AND
  Weizhu Chen \\
  Microsoft Azure AI \\
  \texttt{wzchen@microsoft.com}  \And
  Tuo Zhao \\
  Georgia Institute of Technology \\
  \texttt{tourzhao@gatech.edu} 
  }

\begin{document}
\maketitle

\begin{abstract}


Model ensemble is a popular approach to produce a low-variance and well-generalized model. However, it induces large memory and inference costs, which are often not affordable for real-world deployment. Existing work has resorted to sharing weights among models. However, when increasing the proportion of the shared weights, the resulting models tend to be similar, and the benefits of using model ensemble diminish. To retain ensemble benefits while maintaining a low memory cost, we propose a consistency-regularized ensemble learning approach based on perturbed models, named {\ours}. Specifically, we share the weights of bottom layers across all models and apply different perturbations to the hidden representations for different models, which can effectively promote the model diversity. Meanwhile, we apply a prediction consistency regularizer across the perturbed models to control the variance due to the model diversity. Our experiments using large language models demonstrate that {\ours} significantly improves the generalization performance of the ensemble model. 
Specifically, {\ours} outperforms the standard ensemble of 8 BERT-base models on the GLUE benchmark by $0.7$ with a significantly smaller model size ($114.2$M vs. $880.6$M).

\end{abstract}
\section{Introduction}

Deep Neural Networks (DNNs) have achieved remarkable success in various fields and have become very powerful in learning complicated models \citep{devlin2018bert, brown2020language,he2020deberta}. However, their remarkable representation powers come at the expense of large model variance, which may hurt the model generalization performance. A popular approach for reducing such variance is model ensemble, where the weights or predictions of a set of models are aggregated to produce the predictions \citep{yang2021discussion, dong2020survey}. For example, \citet{zhang2018deep} show that a simple $2$-model ensemble leads to notable improvement over a single model in computer vision tasks.


Despite such notable benefits, model ensemble has not been widely applied to large language models. The major barriers are its enormous storage and expensive inference cost, which linearly scales with the size and the number of models. Therefore, it is often not affordable to ensemble large language models for deployment using memory-constrained and low-latency edge devices.


To alleviate the memory burden, recent works have resorted to a weight-sharing strategy, where all models share the same set of bottom-layer weights, on top of which branches out a set of parallel, un-shared top-layer weights \citep{lan2018knowledge,chen2020online,li2020online}. 
Since the shared weights are optimized to accommodate multiple \textit{diverse} un-shared branches, they can learn shared representations with better generalization \citep{liu2020mtmtdnn, luong2015multi, ruder2019latent}. 

\begin{figure}[htb!]
    \centering
    \includegraphics[width=1.0\linewidth]{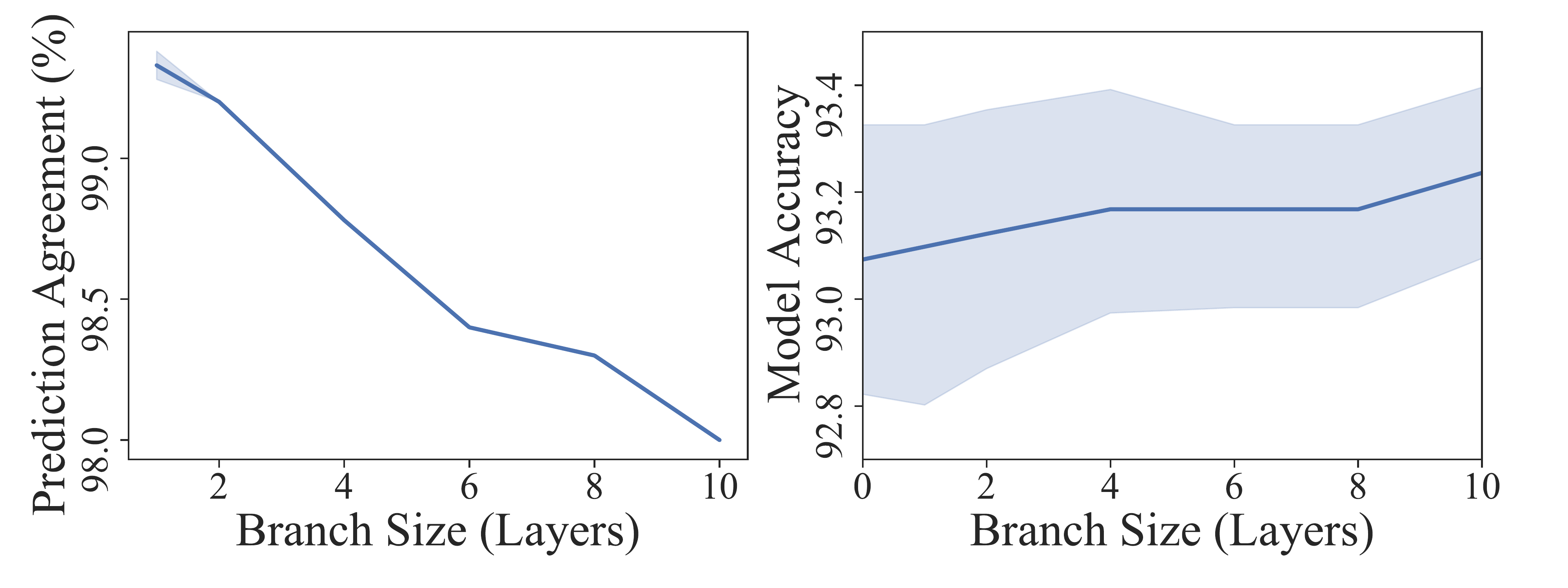}
	\caption{Left: The prediction similarity among branches with different sizes. Right: The average generalization performance and variance of ensembled models over five random seeds. The results are obtained by fine-tuning SST-2 on a BERT-base. A branch size of $0$ corresponds to training a single model.}
	\label{fig:intro}
\end{figure}

For large models, however, such a weight-sharing strategy no longer enjoys the same benefits. Due to memory constraints, a significant proportion of bottom-layer weights need to be shared. Accordingly, top-layer branches have only limited capacity, and therefore the resulting models tend to be similar \citep{chen2020online, rame2021dice, feng2020collaborative, yang2021multi, wu2021peer} (Figure~\ref{fig:intro} (Left)). Due to the lack of the model diversity, their ensemble cannot gain much improvement in generalization. As shown in Figure~\ref{fig:intro} (Right), both the generalization performance and model variance of the ensemble model are similar to those of a single model when the branch size is small. 

To retain a light memory cost while maintaining the ensemble benefits, we propose a new Consistency-regulArized enseMble lEarning approach based on peRturbed mOdels -- {\ours}. Specifically, we share the bottom-layer weights across all models and apply \textit{different} perturbations to the hidden representations for \textit{different} models. Such a perturbation strategy effectively promotes the model diversity. Accordingly, the weights at each layer are optimized to produce consistent outputs given diverse input representations from differently perturbed models. In other words, the shared weights are essentially an on-the-fly ensemble of all perturbed models. In the end, we ensemble all branches on top of the shared weights to produce the final model, which has both a low variance and good generalization performance.


Since we apply perturbations in large models with significant depth, different models' hidden representations may end up being extremely diverse, especially in upper layers. As a result, optimizing the shared weights to accommodate such perturbations can be very challenging. To prevent the models from being over-diverse, we apply a consistency regularizer to reduce variance across different models. Specifically, such a consistency regularizer can be viewed as collaborative distillation across models \citep{guo2020online, lan2018knowledge, zhang2018deep, kim2021feature, chen2020online, li2020online}. By regularizing the discrepancy between each model's output logits and the ensemble of these logits, it encourages all models to be \textit{consistent} in their predictions. We thus adopt consistency regularization to control the perturbed models' diversity from being too large, and thus ease the optimization of the shared weights. 

We conduct thorough experiments to demonstrate {\ours}'s effectiveness and efficiency in ensembling large number of models with more than hundreds of millions of learning parameters. Specifically, our experiments in fine-tuning the BERT-base model on the GLUE benchmark achieve $0.7$ points of gain in terms of task-average score with a significantly smaller parameters over the vanilla ensemble approach ($114.2$M vs. $880.6$M) and achieve $1.2$ points of gain with the same amount of learning parameters over the single model. {\ours} also achieves significant improvements in neural machine translation on both low-resource and high-resource language pairs. 

Furthermore, we verify that {\ours} can learn shared layers with better generalization and ensemble model with smaller variance. We also investigate the effects of using different types and strengths of perturbation and consistency regularization techniques. In particular, we observe that models created with virtual adversarial perturbation \citep{jiang2019smart} and neuron dropout \citep{srivastava2014dropout} lead to ensemble models with the best generalization performance. 
Lastly, we demonstrate {\ours}'s effectiveness on a larger-scale model, RoBERTa-large \citep{liu2019roberta}, where it achieves $0.8$ and $0.9$ points of gain over the vanilla ensemble approach and single model performance, respectively. Our codes are released at \url{https://github.com/cliang1453/CAMERO}.
\section{Background}

\noindent\textbf{Notations.} We use $f(\cdot; \theta)$ to denote a mapping $f$ associated with the parameter $\theta$ from the input sample to an output space, where the output is a multi-dimensional probability simplex for classification tasks and a scalar for regression tasks. We denote the model's final logits as $g(\cdot; \theta)$, where $f(\cdot; \theta) = \sigma(g(\cdot; \theta))$ and $\sigma(\cdot)$ is the Softmax function. We denote $n$ pairs of data samples of the target task as $\{(\boldsymbol{x}_i,y_i)\}_{i=1}^n$. The training loss of $f(\cdot; \theta)$ is computed as $\ell(f(\boldsymbol{x}_i;\theta), y_i)$ for any given training instance $(\boldsymbol{x}_i,y_i)$ where $\ell(\cdot;\cdot)$ denotes the loss function. We use $\mathcal{D}_{KL}(P||Q) = \sum_k p_k \log (p_k / q_k)$ to denote the KL-divergence of two discrete distributions $P$ and $Q$ with the associated parameters of $p_k$'s and $q_k$'s, respectively. 

\noindent\textbf{Collaborative Distillation.} Collaborative distillation approaches train two or more models in parallel while regularizing the consistency of their final prediction distributions \citep{guo2020online, lan2018knowledge, zhang2018deep, kim2021feature, chen2020online, li2020online}. Specifically, we use $\{f(\cdot;\theta_j)\}_{j=1}^m$ to denote $m$ individual models with the same architectures with parameters by $\theta_1, ..., \theta_m$, respectively, and denote $\Theta=\{\theta_1,...,\theta_m\}$. A typical collaborative distillation approach solves the following optimization problem:
\begin{align*}
\min_{\Theta} \mathcal{L}(\Theta) + \alpha \mathcal{R}(\Theta),
\end{align*}
where $\alpha>0$ is a tuning parameter, and $\mathcal{L}(\Theta)$ and $\mathcal{R}(\Theta)$ are defined as
\begin{align}
    &\mathcal{L}(\Theta)= \frac{1}{m}\sum_{j=1}^m \ell(f(\boldsymbol{x};\theta_j), y),\notag\\
    \mathcal{R}(\Theta)&=\frac{1}{m}\sum_{j=1}^m\mathcal{D}(f(\boldsymbol{x}; \theta_j),\mathcal{E}(\boldsymbol{x};\Theta)).
    \label{equ:ensemble-consistency}
\end{align}
For notational simplicity, we will omit the subscript $i$ throughout the rest of the paper. Here, $\mathcal{E}(\boldsymbol{x};\Theta)$ defines a mapping function associated with $\Theta$, which maps the input sample $\boldsymbol{x}$ to a multi-dimensional probability simplex or a scalar depending on the tasks. A commonly adopted ensemble-distillation approach makes $\mathcal{E}(\cdot;\Theta) = \sigma(\sum_{j=1}^m w_j g(\boldsymbol{x};\theta_j))$
, where $\{w_j\}_{j=1}^m$ are non-negative scalars summing to one. $\mathcal{D}(\cdot, \cdot)$ denotes the distance metric of two discrete distributions $P$ and $Q$ or two scalars $p$ and $q$. $\mathcal{D}(P, Q)$ can take the form of symmetric KL-Divergence, $\frac{1}{2}(\mathcal{D}_{KL}(P||Q) + \mathcal{D}_{KL}(Q||P))$. $\mathcal{D}(p, q)$ or euclidean distance $\|p-q\|_2^2$.

\noindent\textbf{Weight-Sharing.} Weight-sharing technique has been adopted in several representation learning scenarios, e.g., multi-task learning \citep{liu2020mtmtdnn, luong2015multi, ruder2019latent}, multi-domain learning \citep{britz2017effective, zeng2018multi, tars2018multi, jiang2019multi} and multi-lingual tasks \citep{gu2018universal, aharoni2019massively}. 
Weight-sharing strategy can reduce the number of free parameters in the model, which helps prevent overfitting in the training and lead to better generalization abilities. 
\section{Method}

\begin{figure}[t!]
    \centering
    \includegraphics[width=1.0\linewidth]{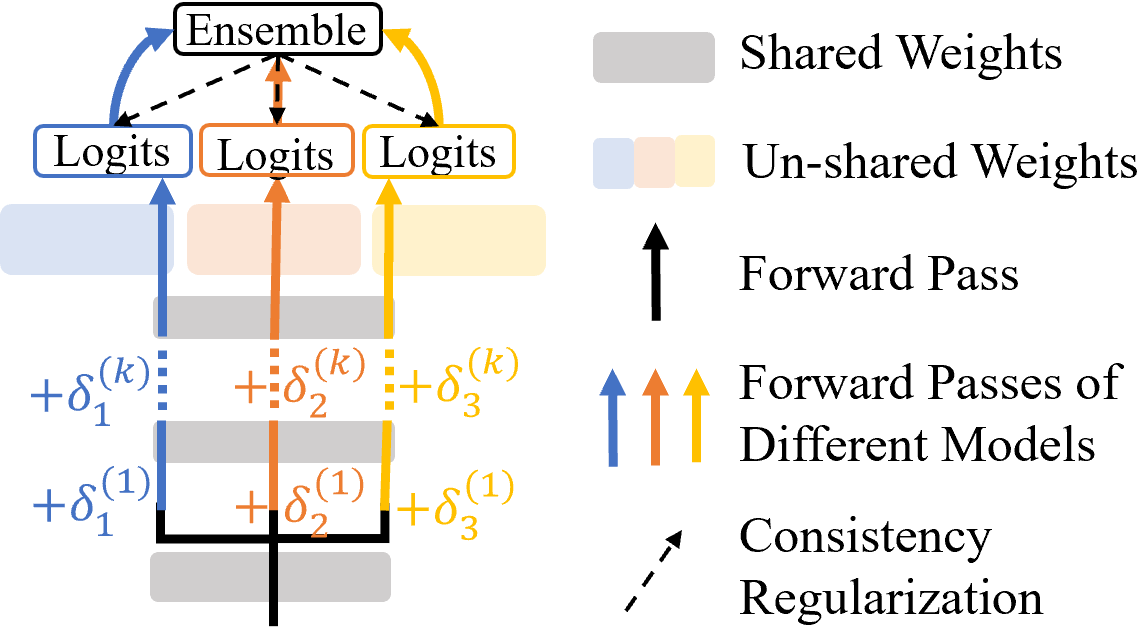}
	\caption{Illustration of CAMERO during training.}
	\label{fig:flowchart}
\end{figure}

We introduce {\ours}, a weight-sharing ensemble learning approach based on consistency-regularized perturbed models.

\subsection{Ensemble Learning w/ Perturbed Models} 
Based on the multi-layer structures of neural networks, we divide each model into two parts: the bottom-layers and the top-layers. The model parameters in bottom-layers are shared across all models. Specifically, the parameters of the $j$-th model is denoted as $\theta_j = [\theta_0,\theta_j^{'}]$, where $\theta_0$ denotes the shared weights in bottom-layers, and $\theta_j^{'}$ denotes the top-layer weights of the $j$-th model. Based on such a compositional structure, the $j$-th model's output can be denoted as
\begin{gather*}
    \begin{split}
        f(\boldsymbol{x};\theta_{j}) = f_{K}(
        f_{K-1}({}&\ldots f_1(\boldsymbol{x};\theta_{j}^{(1)}) \\  &\ldots;  \theta_{j}^{(K-1)}); \theta_{j}^{(K)}), 
    \end{split}
\end{gather*}
where $f_k(\cdot; \theta_j^{(k)})$ is the mapping associated with the $k$-th layer parameter $\theta_j^{(k)}$, and $\theta_0$ consists of $\{\theta_j^{(k)}\}_{k=1}^{K'}$ for some $K'\in\{1,...,K\}$, which shares across models.

A significant proportion of shared weights leads to the models' similarity, which accordingly impairs the ensembled model's performance. To increase the models' diversity, we consider perturbing each layer's hidden representations for different models during training (Figure~\ref{fig:flowchart}). Specifically, the $j$-th model's output is denoted as 
\begin{gather*}
    \begin{split}
        f(\boldsymbol{x};\theta_{j}, {}&\Delta_{j}) = f_{K}(
        f_{K-1}( \ldots f_1(\boldsymbol{x};\theta_{j}^{(1)})
         \\
        & + \delta_{j}^{(1)} \ldots; \theta_{j}^{(K-1)})
        + \delta_{j}^{(K-1)}; \theta_{j}^{(K)}), 
    \end{split}
\end{gather*}
where $\delta_j^{(k)}$ is the perturbation applied at the $k$-th layer's hidden representations, and $\Delta_{j} = \{\delta_{j}^{(k)}\}_{k=1}^{K-1}$, which is sampled from a distribution $\mathcal{P}$. We then train $m$ models with SGD-type algorithms using the following loss:
\begin{gather}
    \begin{split}
        \mathcal{L}_{\Delta}(\Theta) = \frac{1}{m}\sum_{j=1}^m{}&\mathbb{E}_{\Delta_j\sim \mathcal{P}}[\\ &\ell(f(\boldsymbol{x};[\theta_0, \theta_{j}^{'}], \Delta_j), y)].
    \end{split}
    \label{equ:training_loss}
\end{gather}

\begin{remark}
We can consider a wide variety of perturbations for hidden representations, input embeddings or data samples, e.g., random perturbation \citep{aghajanyan2020better}, virtual adversarial perturbation \citep{miyato2018virtual,jiang2019smart}, neuron dropout \citep{srivastava2014dropout} and word dropout \cite{wei2019eda}.
\end{remark}

\subsection{Consistency-Regularized Perturbed Models}

In large models with significant depth, different models' hidden representations may end up being extremely diverse, especially in upper layers. As a result, optimizing the shared weights to accommodate such diverse inputs can be very challenging. To address this issue, we propose to control model variability through consistency regularization. Specifically, we regularize the consistency among $m$ models' final prediction distributions by minimizing the following loss,
\begin{gather}
    \begin{split}
        \mathcal{R}_{\Delta}{}&(\Theta) =\frac{1}{m}\sum_{j=1}^m  \mathbb{E}_{\Delta_j\sim\mathcal{P}}[\\&\mathcal{D}(f(\boldsymbol{x};[\theta_0, \theta_{j}^{'}], \Delta_j), \mathcal{E}(\boldsymbol{x};\Theta, \{\Delta_j\}_{j=1}^m))], \hspace{-0.2in}
    \end{split}
    \label{equ:consistency_loss}
\end{gather}
where $\mathcal{E}(\boldsymbol{x};\Theta, \{\Delta_j\}_{j=1}^m)$ denotes the final prediction distribution produced by some ensemble method applied upon models with perturbed representations. For example, commonly adopted ensemble methods include logits ensemble, where $\mathcal{E}(\boldsymbol{x};\Theta, {\{\Delta_j\}}_{j=1}^{m}) = \sigma(\sum_{j=1}^m w_j g(\boldsymbol{x};\theta_j, \{\Delta_j\}_{j=1}^m))$. In summary, we train $m$ models by minimizing the following overall loss function:
\begin{gather*}
    \mathcal{L}_{\Delta}(\Theta) + \alpha \mathcal{R}_{\Delta}(\Theta),
    \label{equ:our_loss}
\end{gather*}
where $\mathcal{L}_{\Delta}(\Theta)$ is defined in Eq.~(\ref{equ:training_loss}) and $\mathcal{R}_{\Delta}(\Theta)$ is defined in Eq.~(\ref{equ:consistency_loss}). We adjust the strength of consistency regularization via $\alpha$, a non-negative hyper-parameter. 

\begin{remark} 
Different from existing weight-sharing strategies, which control models' diversity via the amount of shared and un-shared weights, {\ours} controls model diversity via the strength of perturbation and regularization. Such a difference renders significant memory benefits. In practice, we safely share all layers except a single top layer. Accordingly, the memory storage is reduced to that of a single model.
This allows us to explore the behaviors of ensemble learning under a larger number of models. 
\end{remark}
\section{Experiment}
We verify the effectiveness of {\ours} on widely used benchmarks for natural language understanding and neural machine translation.

\subsection{Natural Language Understanding}

\newcolumntype{C}{@{\hskip1.5pt}c@{\hskip1.5pt}}
\begin{table*}[htb!]
\centering
\small
\begin{tabular}{C|C|CCCCCCCCCC}
\toprule
\# of & Method & \textbf{MNLI-m/mm} & \textbf{QQP} & \textbf{QNLI} & \textbf{CoLA} & \textbf{SST-2} & \textbf{RTE} & \textbf{MRPC}  & \textbf{STS-B}  & \textbf{Avg.} & \# Param.\\
 Models &  & Acc & Acc/F1 & Acc & Mcc & Acc & Acc & Acc/F1  & P/S Corr & Score & (million)\\\midrule
1&Single & 84.5/84.6 & 91.1/88.1 & 91.2 & 58.7 & 92.9 & 71.1 & 86.2/90.4 & 89.7/89.2 & 83.2 & 109.5 \\\midrule
& Vanilla       & 84.9/85.2 & 91.6/88.7 & 91.8 & 58.2 & 93.2 & 70.6 & 86.2/90.4 & 89.8/89.5 & 83.4 & 220.1 \\ 
& DML           & 85.0/85.5 & 91.6/88.7 & 91.9 & 58.2 & 93.3 & 71.3 & 87.1/90.9 & 89.9/89.5 & 83.6 & 220.1 \\ 
& KDCL          & 85.1/85.6 & 91.7/88.8 & 92.0 & 59.4 & 93.2 & 71.8 & 87.0/90.9 & 89.9/89.5 & 83.8 & 220.1 \\ 
2 & ONE         & 84.5/84.7 & 91.1/88.1 & 91.7 & 59.2 & 93.0 & 70.8 & 87.0/91.1 & 89.7/89.3 & 83.4 & 110.7 \\ \cmidrule{2-12}
& {\ours}       & 85.2/85.7 & 91.6/88.8 & 92.2 & 59.8 & 93.2 & 72.6 & 87.1/90.9 & 89.9/89.5 & 84.0 & 110.7 \\\midrule
& Vanilla       & 85.0/85.2 & 91.7/88.9 & 91.8 & 58.4 & 93.1 & 70.8 & 87.2/91.0 & 90.0/89.6 & 83.5 & 440.3 \\ 
& KDCL          & 85.0/85.7 & 91.7/88.8 & 92.0 & 58.6 & 93.3 & 71.3 & 87.4/91.1 & 90.1/89.6 & 83.7 & 440.3 \\ 
4 & ONE         & 84.6/84.9 & 91.2/88.3 & 91.8 & 58.8 & 93.1 & 71.1 & 87.4/91.1 & 89.8/89.4 & 83.5 & 111.9 \\ \cmidrule{2-12}
& {\ours}       & 85.4/86.1 & 91.8/89.1 & 92.3 & 59.5 & 93.5 & 72.8 & 87.2/91.0 & 90.1/89.7 & 84.2 & 111.9 \\\midrule
&Vanilla        & 85.1/85.5 & 91.7/88.8 & 92.1 & 59.0 & 93.2 & 71.0 & 87.2/91.0 & 90.1/89.7 & 83.7 & 880.6 \\ \cmidrule{2-12}
8& {\ours}      & 85.6/86.3 & 91.9/89.2 & 92.7 & 60.5 & 93.6 & 72.4 & 87.4/91.2 & 90.2/89.8 & 84.4 & 114.2 \\\midrule
\end{tabular}
\caption{Single-task fine-tuning dev results on ensembled BERT-base using the GLUE benchmark. "Single" denotes single model performance. All results are from our own implementation.}
\label{tb:glue-val-bert}
\end{table*}


\begin{table*}[htb!]
\centering
\small
\begin{tabular}{C|C|CCCCCCCCCC}
\toprule
\# of & Method & \textbf{MNLI-m/mm} & \textbf{QQP} & \textbf{QNLI} & \textbf{CoLA} & \textbf{SST-2} & \textbf{RTE} & \textbf{MRPC}  & \textbf{STS-B}  & \textbf{Avg.} & \# Param.\\
Models &    & Acc & Acc/F1 & Acc & Mcc & Acc & Acc & Acc/F1  & P/S Corr & Score & (million) \\\midrule
1 & Single    & 90.2/90.2  & 92.2/-    & 94.7 & 68.0 & 96.4 & 86.6 & 90.9/-    &    -/92.4  & 88.9 & 356.4 \\ \midrule
  & Vanilla   & 90.8/90.5  & 92.4/89.8 & 94.7 & 68.2 & 96.5 & 86.2 & 91.2/93.6 & 92.7/92.5  & 89.0 & 1425.6 \\\cmidrule{2-12}
4 & {\ours}   & 91.1/90.9  & 92.5/90.0 & 95.3 & 70.3 & 97.0 & 87.7 & 91.7/94.0 & 92.8/92.6  & 89.8 & 359.6  \\\bottomrule
\end{tabular}
\caption{Single-task fine-tuning dev results on ensembled RoBERTa-large using the GLUE benchmark. "Single" denotes single model performance from \citet{liu2019roberta}; other results are from our own implementation.}
\label{tb:glue-val-roberta}
\end{table*}

\noindent\textbf{Model and data.} We evaluate the fine-tuning performance of BERT-base (110M) \citep{devlin2018bert} and RoBERTa-large (335M) \citep{liu2019roberta} on the General Language Understanding Evaluation (GLUE,~\citet{wang2018glue}) benchmark. 
GLUE contains nine NLU tasks, including textual entailment, question answering, sentiment analysis, and text similarity. Details about the benchmark are deferred to Appendix~\ref{app:glue-data}.

\noindent\textbf{Baseline methods.} We compare {\ours} with Vanilla, where all models are independently trained without consistency regularization. We also compare {\ours} with representative collaborative distillation methods: Deep Mutual Learning (DML, \citet{zhang2018deep}), On-the-fly Native Ensemble Learning (ONE, \citet{lan2018knowledge}) and Knowledge Distillation via Collaborative Learning (KDCL, \citet{guo2020online})\footnote{We do not include the data augmentation technique proposed in KDCL for a fair comparison.}. DML trains two models with alternating updates while regularizing the consistency between their final prediction distributions. KDCL extends two models to multiple models, training all models concurrently while regularizing the consistency between the prediction distribution of each individual model and of the ensemble of all models. ONE adopts the traditional weight-sharing strategy with a learnable gating factor assigned to each individual branch, which helps to control the model diversity.

\noindent\textbf{Perturbation.} We demonstrate the effectiveness of {\ours} using neuron dropout \citep{srivastava2014dropout}, one of the most straightforward perturbation techniques which randomly zeros out neurons based on a small, fixed ratio. In particular, the ratio adopted in our experiments is $0.1$. In Section~\ref{ana:pert}, we further demonstrate that a wide variety of perturbations, including virtual adversarial perturbation \citep{jiang2019smart}, random perturbation \citep{aghajanyan2020better} and word dropout \citep{wei2019eda}, can all serve the role.

\noindent\textbf{Consistency regularization.} We demonstrate the effectiveness of {\ours} using the ensemble consistency defined in Eq.~(\ref{equ:ensemble-consistency}). In Section~\ref{ana:const}, we further investigate the effectiveness of different types of consistency regularization techniques.

\noindent\textbf{Initialization.} To fine-tune the BERT encoder on downstream tasks, the common initialization approach is to append a randomly initialized, fully connected classification layer on top of the encoder \citep{devlin2018bert}. For ONE and {\ours}, we append $m$ differently initialized, parallel classification layers on top of the encoder. For other methods, we initialize $m$ individual encoders and append a differently initialized classification layer on top of each. 

\noindent\textbf{Inference.} For ONE and {\ours}, we conduct a single pass through the encoder and average the predicted logits of $m$ classification layers. For other methods, we average the predicted logits of $m$ models. All results in the following experiments are evaluated based on such a logits ensemble. 

\noindent\textbf{Implementation details.} Our implementation is based on the MT-DNN code-base\footnote{https://github.com/namisan/mt-dnn}. We follow the suggested training and hyper-parameters settings from \citet{liu2020mtmtdnn}. Specifically, we adopt Adamax \citep{kingma2014adam} as the optimizer with $\beta = (0.9, 0.999)$. We tune $\alpha$ in range of $\{0.5, 1, 2, 5\}$ for all methods. Comprehensive training details are reported in Appendix~\ref{app:glue-training}. 

\noindent\textbf{Results of BERT-base.} Table~\ref{tb:glue-val-bert} shows the evaluation results of BERT-base on the GLUE development set. The results are averaged over five random seeds, and all gains are statistically significant\footnote{All results have passed a paired student t-test with p-values less than $0.05$. The detailed statistics are summarized in Appendix~\ref{app:glue-exp-stats}.}. 

We have the following observations: 1) With significantly less learning parameters, {\ours} achieves a prominent and consistent margin over Vanilla, DML and KDCL. This suggests that {\ours} can produce better-generalized ensemble model with higher parameter efficiency. 2) {\ours} significantly outperforms ONE, suggesting that applying perturbations to models effectively improves the performance of weight-sharing strategy. 3) As the number of models increases from $2$ to $8$, {\ours}'s performance steadily increases for $6$ out of $8$ tasks, while Vanilla and KDCL fail to do so. 

\noindent \textbf{Results of RoBERTa-large.} We further verify that {\ours} can benefit an even larger model, RoBERTa-large. As shown in Table~\ref{tb:glue-val-roberta}, {\ours} achieves consistent gains across all tasks\footnote{We present the median of five runs following \citet{liu2019roberta}.}. Worth noticing, Vanilla shows limited improvements upon the single model performance (e.g., the gains are $0.0$, $0.1$ and $-0.4$ on QNLI, SST-2 and RTE, respectively). We conjecture that the high model variance in large models compromises the ensemble benefits. In contrast, by control the model variance with regularization, {\ours} achieves gains of $0.6$, $0.6$ and $1.1$ on these tasks.

\subsection{Neural Machine Translation}

\newcolumntype{C}{@{\hskip4pt}c@{\hskip4pt}}
\begin{table*}[htb!]
\centering
\small
\begin{tabular}{C|C|CCCCC|CCCCCC}
\toprule
\# of  & Method & \multicolumn{5}{C}{\textbf{IWSLT}} & \multicolumn{4}{C}{\textbf{WMT}} \\
Models &  & \textbf{En-De} & \textbf{De-En} & \textbf{En-Fr} & \textbf{Fr-En} & \textbf{Avg.} & \textbf{En-De}& \textbf{De-En} & \textbf{Avg.} & \# Param. \\\midrule
1 & Single& 28.5 & 34.7 & 38.1 & 37.7 & 34.7 & 26.9 & 30.7 & 28.8 & 54.5 \\ \midrule
& Vanilla & 28.6 & 34.8 & 38.2 & 37.8 & 34.9 & 27.0 & 31.2 & 29.1 & 109.1  \\ 
& DML     & 30.5 & 37.4 & 39.9 & 39.6 & 36.9 & 27.1 & 31.8 & 29.5 & 109.1   \\ 
& KDCL    & 30.6 & 37.2 & 39.8 & 39.5 & 36.7 & 27.2 & 31.9 & 29.6 & 109.1   \\ 
2 & ONE   & 28.9 & 35.1 & 38.5 & 38.2 & 35.2 & 27.0 & 31.0 & 29.0 & 58.7 \\\cmidrule{2-11}
& {\ours} & 30.8 & 37.5 & 40.2 & 39.8 & 37.1 & 27.6 & 32.2 & 29.9 & 58.7  \\\midrule
& Vanilla & 28.7 & 34.9 & 38.2 & 37.8 & 34.9 & 27.0 & 31.2 & 29.1 & 218.1  \\ 
& KDCL    & 30.8 & 37.4 & 39.9 & 39.7 & 36.9 & 27.1 & 32.0 & 29.6 & 218.1  \\    
4 & ONE   & 28.8 & 35.0 & 38.2 & 37.9 & 35.0 & 27.1 & 31.1 & 29.1 & 67.1   \\\cmidrule{2-11}
& {\ours} & 31.1 & 37.8 & 40.3 & 39.9 & 37.3 & 27.7 & 32.4 & 30.1 & 67.1  \\\bottomrule
\end{tabular}
\caption{Test set scores on ensembled Transformer-base on IWSLT tasks (BLEU) and WMT tasks (SacreBLEU). "Single" denotes single model performance. All results are from our own implementation.}
\label{tb:nmt-val}
\end{table*}

\noindent\textbf{Model and data.} We further evaluate {\ours} on the Transformer-base NMT model \citep{vaswani2017attention} using widely used IWSLT \citep{cettolo2016iwslt}\footnote{https://wit3.fbk.eu/} and WMT \citep{bojar2016findings}\footnote{http://data.statmt.org/wmt16/translation-task/} datasets. Specifically, we adopt IWSLT'14 En$\leftrightarrow$De, IWSLT'16 En$\leftrightarrow$Fr and WMT'14 En$\leftrightarrow$De. IWSLT En$\leftrightarrow$De and En$\leftrightarrow$Fr are low-resource datasets containing 160k and 236k sentence pairs. WMT En$\leftrightarrow$De is a rich-resource dataset containing 4.5M sentence pairs. Model and dataset details are deferred to Appendix~\ref{app:nmt-data}.

\noindent\textbf{Implementation details.} Our implementation is based on the fairseq code-base and follows the training and hyper-parameters settings from \citet{ott2018scaling, ott2019fairseq}. Specifically, we use $5\times10^{-4}$ as the learning rate and employ Adam \citep{kingma2014adam} as the optimizer with $\beta=(0.9, 0.98)$. We select $\alpha$ in range of $\{1,2,5\}$. For ONE and {\ours}, we randomly initialize multiple parallel decoders's last layers as the un-shared branches. 
Comprehensive training details are reported in Appendix~\ref{app:nmt-training}. 

\noindent\textbf{Main results.} Table~\ref{tb:nmt-val} shows the BLEU scores on the IWSLT test set and the SacreBLEU scores \citep{post2018call} with compound splitting on the WMT test set\footnote{We evaluate the SacreBLEU score on the average of last $10$ checkpoints. The tokenizer version is: nrefs:1 | case:mixed | eff:no | tok:13a, smooth:exp | version:2.0.0.}. WMT's corresponding BLEU scores are reported in Appendix~\ref{app:nmt-exp}. 

With a number of learning parameters similar to a single model, {\ours} achieves around $2$ and $1$ points upon ONE, and improves around $0.4$ and $0.4$ points upon KDCL, on low-resource and rich-resource datasets, respectively. This suggests that other than fine-tuning, {\ours} also improves the generalization of training-from-scratch models in both low-resource and rich-resource datasets.

\section{Analysis}

We first verify that {\ours} leads to a well-generalized and low-variance ensemble model. We then demonstrate how the perturbation and consistency regularization strength influences the model diversity and performance. Finally, we demonstrate {\ours}'s effectiveness on various types of perturbation and regularization techniques. 

\subsection{Shared Weights Learn Better Representations} 

We verify that {\ours} allows the shared weights to learn better-generalized representations. Specifically, we attach a randomly initialized classifier on top of a BERT-base encoder trained by {\ours}. We then fix the encoder and fine-tune the attached classifier only. As shown in Table~\ref{tb:glue-val-n-students}, the encoder trained by {\ours} learns better representations than ONE's consistently across different tasks and under different numbers of models.

\begin{table}[htb!]
\centering
\small
\begin{tabular}{C|C|CCCCC}
\toprule
\# of      & Methods & \textbf{MNLI} & \textbf{SST-2} & \textbf{MRPC}  & \textbf{Avg.} \\
Models    & & Acc                & Acc            & Acc/F1       & Score        \\\midrule
1           & Single & 84.58              & 92.95          & 86.88          & 88.14         \\ \midrule
2           & ONE    & 84.53              & 92.94          & 88.94          & 88.80          \\
            & {\ours}& 85.41              & 93.03          & 89.04          & 89.16          \\ \midrule
4           & ONE    & 84.67              & 93.03          & 89.07          & 88.92          \\
            & {\ours}& 85.57              & 93.46          & 89.20          & 89.41          \\ \bottomrule
\end{tabular}
\caption{Performance of the ensembled BERT-base encoder using the GLUE dev set. We only fine-tune the randomly initialized classification layer on top of a well-trained encoder.}
\label{tb:glue-val-n-students}
\end{table}

\subsection{Ensemble Model Has a Low Variance Across Random Seeds} 

\begin{figure}[htb!]
    \centering
    \includegraphics[width=1.0\linewidth]{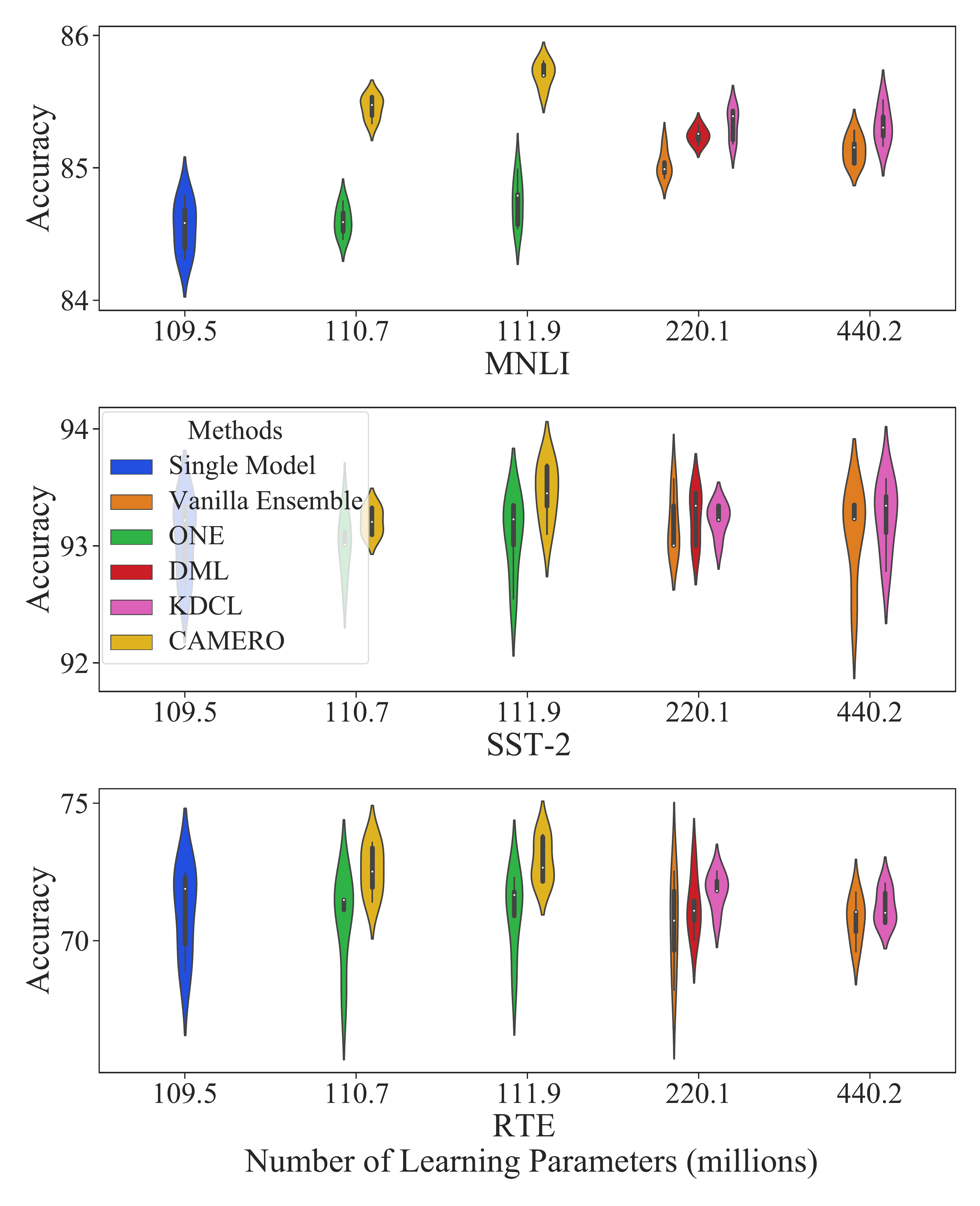}
	\caption{Performance and variance of the ensembled BERT-base on the GLUE dev set.}
	\label{fig:main_claim}
\end{figure}

We verify that {\ours} produces an ensemble model that both generalizes well and has a low variance across different random seeds under a light parameter budget. Figure~\ref{fig:main_claim} plots the prediction accuracy of $2$-model and $4$-model ensemble across five seeds. For example, in MNLI, {\ours}'s $2$-model ensemble ($110.7$M) achieves similar performance to KDCL's $4$-model ensemble (440.2M) and {\ours}'s $4$-model ensemble (111.9M) achieves an even better performance. Across different tasks, {\ours}'s ensemble model has a similar or lower variance than all others. Complete variance statistics are presented in Appendix~\ref{app:glue-exp-stats}. 

\begin{table*}[htb!]
\centering
\small
\begin{tabular}{l|CCCCCC}
\toprule
Perturbation Types       & \textbf{MNLI} & \textbf{QNLI} & \textbf{SST-2}  & \textbf{MRPC}  & \textbf{CoLA}  &  \textbf{Avg.} \\
                         & Acc                & Acc           & Acc             & Acc/F1         & MCC            & Score \\\midrule
None                      & 84.74 & 91.76 & 93.10 & 89.05 & 58.75 & 83.48 \\ 
Neuron Dropout            \citep{srivastava2014dropout} & 85.73 & 92.30 & 93.46 & 89.09 & 59.50 & 84.02 \\
Virtual Adversarial Pert. \citep{jiang2019smart} & 85.76 & 92.33 & 93.53 & 89.19 & 59.49 & 84.08 \\
Random Noise Pert.        \citep{aghajanyan2020better} & 85.78 & 92.21 & 93.42 & 89.07 & 59.22 & 83.94 \\ 
Word Dropout              \cite{wei2019eda} & 85.61 & 92.00 & 93.21 & 89.06 & 59.19 & 83.81 \\ \bottomrule
\end{tabular}
\caption{{\ours}'s performance under different types of perturbation. "None" corresponds to ONE, which does not apply different perturbations to different models. We report the $4$-model ensembled BERT-base results. }
\label{tb:perturbation}
\end{table*}

\begin{figure*}[htb!]
    \centering
    \includegraphics[width=0.7\linewidth]{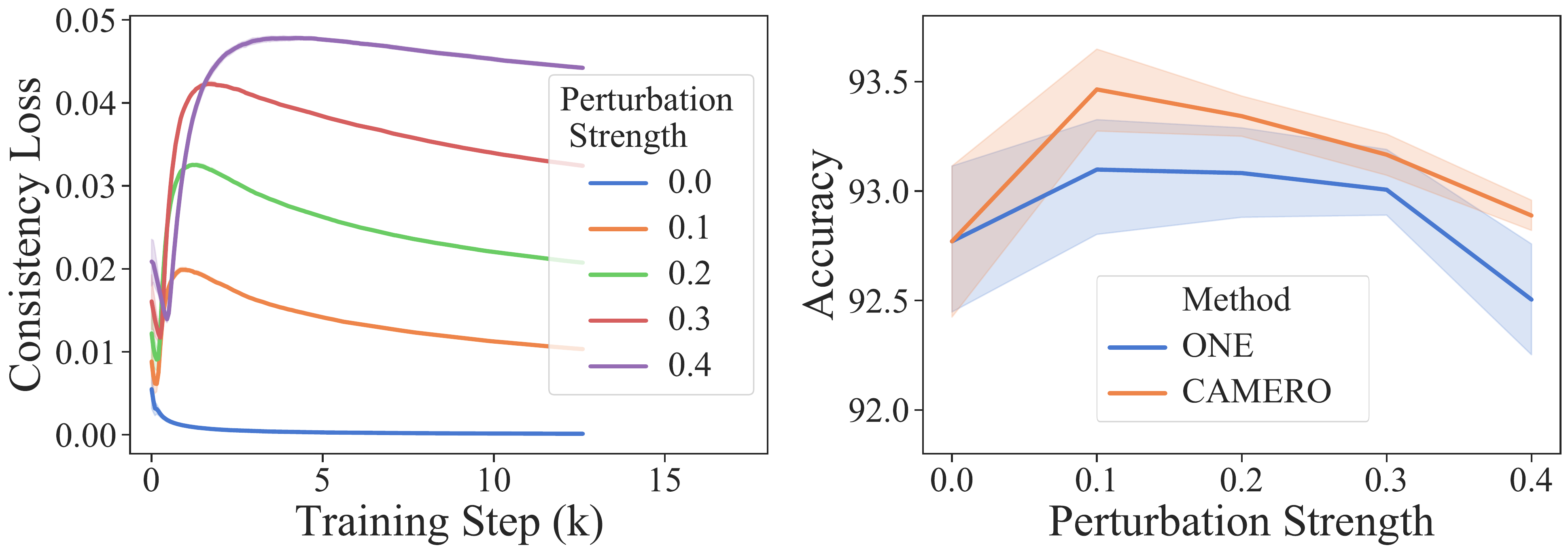}
	\caption{The effect of perturbation strength on models' diversity during training (Left) and the variance of the ensembled model (Right). We fine-tune BERT-base on SST-2 and report the $4$-model ensemble results. }
	\label{fig:perturbation}
\end{figure*}

\subsection{Types and Strength of Perturbations}
\label{ana:pert}

\begin{table*}[htb!]
\centering
\small
\begin{tabular}{l|CCCCCC}
\toprule
Consistency Types    & \textbf{MNLI} & \textbf{QNLI} & \textbf{SST-2}  & \textbf{MRPC}  & \textbf{CoLA}  &  \textbf{Avg.} \\
                      & Acc                & Acc           & Acc             & Acc/F1         & MCC            & Score \\\midrule
None                      & 85.23 & 91.76 & 93.30 & 88.97 & 58.40 & 83.48 \\ 
Ensemble Consistency          & 85.73 & 92.30 & 93.46 & 89.09 & 59.50 & 84.02 \\
Pairwise Consistency          & 85.73 & 92.33 & 93.37 & 89.40 & 59.87 & 84.14 \\
\bottomrule
\end{tabular}
\caption{{\ours}'s performance under different types of consistency regularization. "None" corresponds to $\alpha=0$, where no regularization is applied. We report the $4$-model ensembled BERT-base results.}
\label{tb:consistency}
\end{table*}

\begin{figure*}[htb!]
    \centering
    \includegraphics[width=0.85\linewidth]{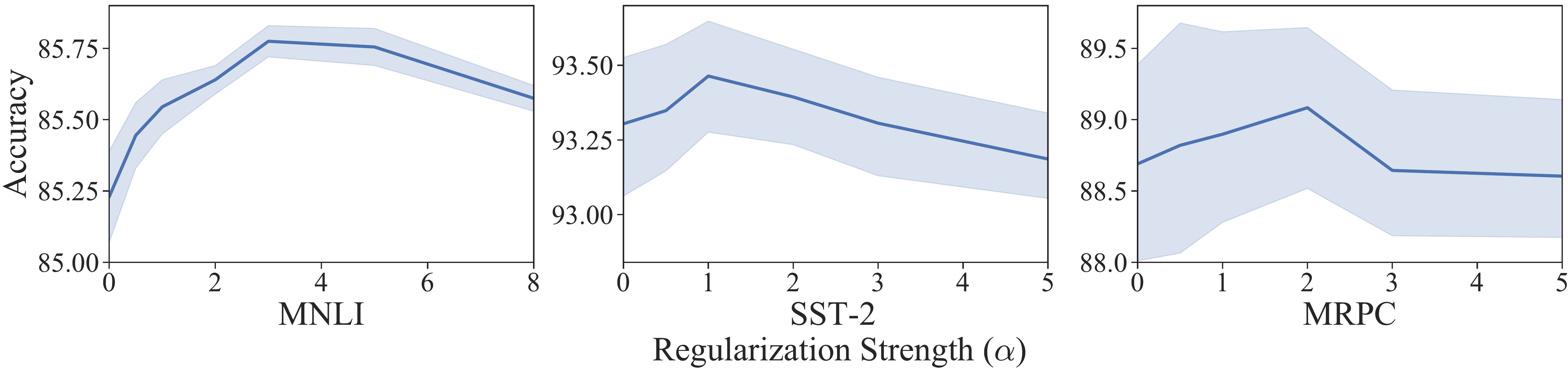}
	\caption{The effect of consistency regularization strength on the generalization and variance of the ensembled model. We fine-tune BERT-base and report the $4$-model ensemble results.}
	\label{fig:consistency}
\end{figure*}

\noindent\textbf{Types of perturbation.} We verify that {\ours} produces well-generalized ensemble models under various types of perturbations. Specifically, we apply virtual adversarial perturbation \citep{jiang2019smart} and random noise perturbation \citep{aghajanyan2020better} on the first layer input embeddings, neuron dropout on all layers' input representations \citep{srivastava2014dropout}, and word dropout on the input sentences \cite{wei2019eda}. Specifically, we set the dropout ratio to be $0.1$ for neuron dropout and $0.05$ for word dropout. We set the norm constraint $\epsilon = 1\times 10^{-5}$ for both virtual adversarial perturbation and random noise perturbation. The random noise is sampled from a normal distribution. As shown in Table~\ref{tb:perturbation}, {\ours} leads to significant margin of improvement under all types of perturbation. In particular, virtual adversarial perturbation and neuron dropout perform consistently well on all tasks. Random noise perturbation performs well on larger tasks (e.g., MNLI, QNLI, SST-2) while the gains shrink on smaller tasks.

\noindent\textbf{Strength of perturbation.} We then verify that a larger perturbation strength improves the perturbed models' diversity during training. As consistency loss is computed as the average distance between all perturbed models' output logits to the ensembled logits at each iteration, it directly reflects the model diversity during training. As shown in Figure~\ref{fig:perturbation} (Left), a larger neuron dropout ratio leads to larger consistency loss, thus higher model diversity. 

Furthermore, we observe that a larger perturbation strength leads to a lower-variance ensemble model. As shown in Figure~\ref{fig:perturbation} (Right), as the neuron dropout ratio grows, {\ours}'s ensemble model variance decreases. In contrast, ONE has a large variance under all ratios. 

\subsection{Types and Strength of Consistency Regularization}
\label{ana:const}

\noindent\textbf{Types of consistency regularization.} We then investigate the effects of using different types of consistency regularization techniques. Specifically, we compare the existing \textit{ensemble consistency}, as defined in Eq.~(\ref{equ:ensemble-consistency}), and a newly proposed \textit{pairwise consistency}, which is defined as
\begin{gather*}
\begin{split}
    \mathcal{R}(\Theta) = \frac{2}{m(m-1)}\sum_{j=1}^m\sum_{p=j+1}^m \mathcal{D}({}&f(\boldsymbol{x}; \theta_j),\\ & f(\boldsymbol{x}; \theta_p)).
\end{split}
\label{equ:pairwise-consistency}
\end{gather*}
The pairwise consistency measures the average distance between \textit{each pair} of models' output logits, thus we expect it to capture the discrepancies among models more accurately. 
As shown in Table~\ref{tb:consistency}, {\ours} shows consistent improvements under both types of regularization. In particular, pairwise consistency shows larger advantages on smaller tasks (e.g., $0.3$ on MRPC and $0.4$ on CoLA).  

\noindent\textbf{Strength of consistency regularization.} We further investigate how the strength of the regularization factor $\alpha$ affects the ensemble model's performance. As shown in Figure~\ref{fig:consistency}, as $\alpha$ increases, the generalization performance of the ensemble model first increases, then decreases. This suggests that regularization can effectively benefits the generalization performance through balancing the model diversity. 
\section{Conclusion}

We propose {\ours}, a consistency-regularized ensemble learning approach based on perturbed models. Such a strategy significantly improves the parameter efficiency of model ensemble in large language models, making it an accessible and powerful technique for learning ensemble models with better generalization performances.

\clearpage
\bibliography{anthology,custom}
\bibliographystyle{acl_natbib}

\appendix
\section{Appendix}
\label{appendix}

\subsection{Natural Language Understanding}

\subsubsection{Data}
\label{app:glue-data}

GLUE is a collection of nine NLU tasks. The benchmark includes question answering \citep{squad1}, linguistic acceptability (CoLA, \citealt{cola2018}), sentiment analysis (SST, \citealt{sst2013}), text similarity (STS-B, \citealt{sts-b2017}), paraphrase detection (MRPC, \citealt{mrpc2005}), and natural language inference (RTE \& MNLI, \citealt{rte1,rte2,rte3,rte5,mnli2018}) tasks. Details of the GLUE benchmark, including tasks, statistics, and evaluation metrics, are summarized in Table~\ref{tab:glue}. 

All the texts were tokenized using wordpieces, and were chopped to spans no longer than $512$ tokens.

\subsubsection{Training Details}
\label{app:glue-training}

Table~\ref{tb:glue-hyperparam} presents the hyper-parameter configurations to fine-tune BERT-base and RoBERTa-large models. We apply a linear weight decay rate of $0.01$ and a gradient norm clipping threshold of $1$ for all experiments. All experiments are conducted on Nvidia V100 GPUs.

\newcolumntype{C}{@{\hskip3pt}c@{\hskip3pt}}
\begin{table*}[htb!]
\centering
\footnotesize
\begin{tabular}{l|l|CCCCCCCCC}
\toprule
\textbf{Hyper-param} & \textbf{Model}                      &\textbf{RTE} & \textbf{MRPC} & \textbf{CoLA} & \textbf{SST-2} & \textbf{STS-B} & \textbf{QNLI} & \textbf{QQP} & \textbf{MNLI} \\ \midrule
Learning Rate & BERT\textsubscript{BASE} & 1e-4 & 1e-4 & 1e-4 & 8e-5 & 1e-4 & 1e-4 & 1e-4 & 8e-5 \\
& RoBERTa\textsubscript{LARGE}           & 5e-5 & 1e-4 & 3e-5 & 2e-5 & 5e-5 & 1e-5 & 1e-4 & 3e-5 \\\midrule
Epoch & BERT\textsubscript{BASE}         & 6 & 6 & 6 & 6 & 6 & 3 & 6 & 3 \\
& RoBERTa\textsubscript{LARGE}           & 15 & 6 & 6 & 10 & 10 & 10 & 10 & 3  \\\midrule
Batch Size & BERT\textsubscript{BASE}    & 16 & 8 & 32 & 32 & 32 & 32 & 32 & 32\\
& RoBERTa\textsubscript{LARGE}           & 8 & 16 & 32 & 32 & 32 & 32 & 32 & 32\\\midrule
Dropout & Both & 0.1 & 0.1 & 0.1 & 0.1 & 0.1 & 0.1 & 0.1 & 0.3 \\\midrule
Warmup & BERT\textsubscript{BASE} & 0.1 & 0.1 & 0.1 & 0.1 & 0.1 & 0.1 & 0.1 & 0.1  \\
\bottomrule
\end{tabular}
\caption{Hyper-parameter configurations for GLUE experiments. ``Epoch'' refers to the total training epochs; we adopt early-stopping strategy in practice. ``Dropout'' refers to classification layer dropout ratio, the encoder dropout ratio is fixed to be $0.1$. ``Warmup'' refers to the ratio of learning rate linear warmup iterations to total training iterations. }
\label{tb:glue-hyperparam}
\end{table*}

\subsubsection{Evaluation Results}
\label{app:glue-exp-stats}
\textbf{Statistics of the dev set results.} Table~\ref{tb:glue-dev-stats} shows the standard deviation of the dev set results.
\newcolumntype{C}{@{\hskip3pt}c@{\hskip3pt}}
\begin{table*}[htb!]
\centering
\small
\begin{tabular}{C|CCCCCCCCC}
\toprule
\# of & Method & \textbf{MNLI-m/mm} & \textbf{QQP} & \textbf{QNLI} & \textbf{CoLA} & \textbf{SST-2} & \textbf{RTE} & \textbf{MRPC}  & \textbf{STS-B}  \\
 Models &  & Acc & Acc/F1 & Acc & Mcc & Acc & Acc & Acc/F1  & P/S Corr \\\midrule
1 & Single      & 0.20      & 0.25      & 0.21 & 1.10 & 0.33 & 1.61 & 0.83      & 0.20      \\ \midrule
& Vanilla       & 0.17      & 0.11      & 0.16 & 1.22 & 0.26 & 1.72 & 0.77      & 0.21      \\ 
& DML           & 0.14      & 0.03      & 0.15 & 0.76 & 0.23 & 1.07 & 0.69      & 0.19      \\ 
& KDCL          & 0.11      & 0.05      & 0.16 & 1.04 & 0.14 & 0.68 & 0.66      & 0.19      \\ 
2 & ONE         & 0.13      & 0.22      & 0.21 & 1.00 & 0.25 & 1.61 & 0.88      & 0.15      \\ \cmidrule{2-10}
& {\ours}       & 0.11      & 0.05      & 0.15 & 1.01 & 0.12 & 0.92 & 0.37      & 0.11      \\ \midrule
& Vanilla       & 0.14      & 0.10      & 0.09 & 0.92 & 0.39 & 0.82 & 0.54      & 0.11      \\ 
& KDCL          & 0.20      & 0.12      & 0.12 & 1.28 & 0.31 & 0.66 & 0.27      & 0.06      \\  
4 & ONE         & 0.19      & 0.22      & 0.25 & 0.98 & 0.34 & 1.44 & 0.84      & 0.12      \\  \cmidrule{2-10}
& {\ours}       & 0.11      & 0.07      & 0.05 & 1.03 & 0.25 & 0.84 & 0.19      & 0.06       \\ \bottomrule
\end{tabular}
\caption{Standard deviation of the single-task fine-tuning dev results on ensembled BERT-base.}
\label{tb:glue-dev-stats}
\end{table*}

\noindent \textbf{Average score computation formula.} For dev set results, we first obtain a score for each task by averaging the scores of all metrics (e.g., Acc and F1) and test sets (e.g., MNLI-m and MNLI-mm) within this task, then compute a task-average score. For test set results, we directly averages scores of all reported metrics following \citet{devlin2018bert}.

\subsection{Neural Machine Translation}

\subsubsection{Data}
\label{app:nmt-data}

For IWSLT'14 En-De and De-En datasets, we follow \citet{ott2019fairseq}\footnote{https://github.com/pytorch/fairseq/blob/main/examples\\/translation/prepare-iwslt14.sh} to split the train/dev/test set. For IWSLT'16 En-Fr and Fr-En, we adopt the default training set, and use IWSLT16.TED.tst2015 for validation and use IWSLT16.TED.tst2016 for testing. For WMT'14 En-De and De-En, We use the standard newstest-2013 and newstest-2014 for validation and testing, respectively. Table~\ref{tab:nmt-data} shows the number of sentence pairs in each dataset.

\newcolumntype{C}{@{\hskip3pt}c@{\hskip3pt}}
\begin{table*}[!htb]
\centering
\begin{tabular}{l|CCC} \toprule 
\textbf{Data}  &  \textbf{Train} & \textbf{Dev} & \textbf{Test} \\ \midrule
\textbf{IWSLT'14 En-De/De-En} &  160  & 7283  & 6750 \\
\textbf{IWSLT'16 En-Fr/Fr-En} &  218  & 1080  & 1133 \\
\textbf{WMT'14 En-De/De-En}   &  4.5m & 1061  & 1019 \\
\bottomrule
\end{tabular}
\caption{The number of parallel sentences in NMT datasets.}
\label{tab:nmt-data}
\end{table*} 

We tokenize all datasets with byte-pair encoding (BPE, \citet{sennrich2015neural}) with a vocabulary size of 10k for datasets in IWSLT and 32k for datasets in WMT. We build a joint dictionary upon all source and target sentences for all datasets. 

\subsubsection{Training Details}
\label{app:nmt-training}

We adopt the Transformer-base model for all datasets and share all embeddings. For IWSLT datasets, we follow the training configurations from \citet{ott2019fairseq}\footnote{https://github.com/pytorch/fairseq/tree/main/examples\\/translation\#iwslt14-german-to-english-transformer}. For WMT datasets, we follow the training configurations from \citet{ott2018scaling}\footnote{https://github.com/pytorch/fairseq/tree/main/examples\\/scaling\_nmt\#training-a-new-model-on-wmt16-en-de}. For all datasets, we use Adam\citep{kingma2014adam} as the optimizer with $\beta=(0.9,0.98)$. We use a inverse square root learning rate schedule. We apply a linear weight decay rate of $1\times10^{-4}$ and a label smoothing ratio of $0.1$ for all experiments. All experiments are conducted on Nvidia V100 GPUs. Table~\ref{tb:nmt-hyperparam} presents the training hyper-parameter configurations for all datasets. 

For evaluation on IWSLT datasets, we report the BLEU score of the best checkpoint using a beam size of $5$ and length penalty of $1$. For evaluation on WMT datasets, we average the last $10$ checkpoints, decode with a beam size of $4$ and length penalty of $0.6$, then report the SacreBLEU scores after compound splitting.

\newcolumntype{C}{@{\hskip3pt}c@{\hskip3pt}}
\begin{table*}[htb!]
\centering
\small
\begin{tabular}{l|CC}
\toprule
\textbf{Hyper-param} & \textbf{IWSLT} & \textbf{WMT} \\ \midrule
Learning Rate & $5\times 10^{-4}$ & $1\times 10^{-3}$  \\\midrule
Batch size & 4096/GPU $\times$ 1 GPU & 3584/GPU$\times$ 8 GPUs $\times$ 16 grad. acc. steps \\\midrule
Epoch & 250 & 150 \\\midrule
Dropout & 0.3 & 0.1 \\\midrule
Warmup & 8000 & 4000 \\\bottomrule
\end{tabular}
\caption{Hyper-parameter configurations for NMT experiments. ``Warmup'' refers to the learning rate linear warmup iterations.}
\label{tb:nmt-hyperparam}
\end{table*}

\subsubsection{BLEU scores for WMT experiments}
\label{app:nmt-exp}
Table~\ref{tb:nmt-bleu} shows the corresponding BLEU scores for WMT datasets. 

\newcolumntype{C}{@{\hskip4pt}c@{\hskip4pt}}
\begin{table*}[htb!]
\centering
\small
\begin{tabular}{C|C|CC}
\toprule
\# of  & Method & \multicolumn{2}{C}{\textbf{WMT}} \\
Models &  & \textbf{En-De} & \textbf{De-En} \\\midrule
  & Single \citep{vaswani2017attention}& 27.30 & - \\
1 & Single& 27.54 & 31.28  \\ \midrule
& Vanilla & 27.62 & 31.76   \\ 
& DML     & 27.70 & 32.22    \\ 
& KDCL    & 27.84 & 32.35   \\ 
2 & ONE   & 27.68 & 31.43   \\\cmidrule{2-4}
& {\ours} & 28.26 & 32.61   \\\midrule
& Vanilla & 27.67 & 31.79   \\ 
& KDCL    & 27.75 & 32.47 \\    
4 & ONE   & 27.78 & 31.48  \\\cmidrule{2-4}
& {\ours} & 28.43 & 32.78 \\\bottomrule
\end{tabular}
\caption{Test set scores on ensembled Transformer-base on WMT tasks (BLEU). The result in "Single \citep{vaswani2017attention}" is the single model performance reported from \citet{vaswani2017attention}; Other results are from our own implementation.}
\label{tb:nmt-bleu}
\end{table*}

\begin{table*}[htb]
	\begin{center}
		\begin{tabular}{l|l|c|c|c|c|c}
			\toprule 
			\bf Corpus &Task& \#Train & \#Dev & \#Test   & \#Label &Metrics\\ \midrule
			\multicolumn{6}{@{\hskip1pt}r@{\hskip1pt}}{Single-Sentence Classification (GLUE)} \\ \hline
			CoLA & Acceptability&8.5k & 1k & 1k & 2 & Matthews corr\\ \hline
			SST & Sentiment&67k & 872 & 1.8k & 2 & Accuracy\\ \midrule
			\multicolumn{6}{@{\hskip1pt}r@{\hskip1pt}}{Pairwise Text Classification (GLUE)} \\ \hline
			MNLI & NLI& 393k& 20k & 20k& 3 & Accuracy\\ \hline
            RTE & NLI &2.5k & 276 & 3k & 2 & Accuracy \\ \hline
			QQP & Paraphrase&364k & 40k & 391k& 2 & Accuracy/F1\\ \hline
            MRPC & Paraphrase &3.7k & 408 & 1.7k& 2&Accuracy/F1\\ \hline
			QNLI & QA/NLI& 108k &5.7k&5.7k&2& Accuracy\\ \midrule
			\multicolumn{6}{@{\hskip1pt}r@{\hskip1pt}}{Text Similarity (GLUE)} \\ \hline
			STS-B & Similarity &7k &1.5k& 1.4k &1 & Pearson/Spearman corr\\ \bottomrule
		\end{tabular}
	\end{center}
	\caption{Summary of the GLUE benchmark.}
	\label{tab:glue}
\end{table*}

\end{document}